\documentclass[conference]{IEEEtran}
\IEEEoverridecommandlockouts
\usepackage{cite}
\usepackage{amsmath,amssymb,amsfonts}

\usepackage{cite}
\usepackage{amsmath,amssymb,amsfonts}
\usepackage{algorithmic}
\usepackage{graphicx}
\usepackage{textcomp}
\usepackage{xcolor}
\usepackage{algorithmic}
\usepackage{graphicx}
\usepackage{textcomp}
\usepackage{xcolor}
\usepackage{graphicx}
\graphicspath{ {./ }}
\def\BibTeX{{\rm B\kern-.05em{\sc i\kern-.025em b}\kern-.08em
    T\kern-.1667em\lower.7ex\hbox{E}\kern-.125emX}}
\begin{document}
\title{Eye Gaze Estimation Model Analysis\\
}

\author{
\IEEEauthorblockN{Aveena Kottwani}
\IEEEauthorblockA{\textit{Department of Computer Science} \\
\textit{Stony Brook University, NY, USA}\\
aveena.kottwani@stonybrook.edu}

\and

\IEEEauthorblockN{Ayush Kumar}
\IEEEauthorblockA{\textit{Department of Computer Science} \\
\textit{Stony Brook University, NY, USA}\\
aykumar@cs.stonybrook.edu}


}

\maketitle
\begin{abstract}
We explore techniques for eye gaze estimation using machine learning. Eye gaze estimation is a common problem for various behavior analysis and human-computer interfaces. The purpose of this work is to discuss various model types for eye gaze estimation and present the results from predicting gaze direction using eye landmarks in unconstrained settings.  In unconstrained real-world settings,  feature-based and model-based methods are outperformed by recent appearance-based methods due to factors like illumination changes and other visual artifacts. We discuss a learning-based method for eye region landmark localization trained exclusively on synthetic data. We discuss how to use detected landmarks as input to iterative model-fitting and lightweight learning-based gaze estimation methods and how to use the model for person-independent and personalized gaze estimations.
\end{abstract}


\section{Introduction}

Human gaze direction can assist users with motor disabilities using eye-gaze tracking cursor control system, gaze-based human-computer interaction , visual attention analysis~\cite{vcompare}, consumer behavior research, next-generation UIs, marketing analysis, AR, VR and automation of vehicles. Eye-gaze estimation using web cameras can be used by the  users with motor-disabilities in everyday tasks~\cite{task}, such as reading~\cite{spatme}, gaze-based interaction ,digital signage and human-computer interaction. Gaze tracking is the process of measuring either the point of gaze or the motion of an eye relative to the head. Accurate eye gaze
tracking models requires a manual calibration procedure for each new user or expensive specialized hardware. This does not appeal consumer market applications. Since prices of cameras has decreased, this encourages the integration of gaze estimation function in consumer-grade devices equipped with monocular cameras for applications such next-generation controllers for games~\cite{agames}, natural user interfaces, and human–computer interaction (HCI)~\cite{kumarETRA, kumarJEMR}. There is a need for creating gaze estimation function that uses consumer-grade devices equipped with monocular cameras under conditions like low-resolution eye images, unconstrained user postures, different locations, and varying distances from the camera. Recently, inexpensive solutions that work at lower cost and complexity and require no active illumination have been proposed \cite{b1}. Most of these systems are built using modern computer vision algorithms that work with camera in any computer screen, without additional hardware. 

Traditional feature-based and model-based gaze estimation typically rely on accurate detection of eye region landmarks, such
as the iris center or the eye corners. Many previous works have therefore focused on accurately localizing iris center and eye corners landmarks.Using large-scale data-sets we can now have user-specific calibration-free point of gaze (PoG) estimation with a monocular camera.But, the head positions in conventional data sets are constrained considering only the region near to the camera which cannot be used for images taken from more varied distances. In addition, the requirement of high-resolution eye images also limits the practical applicability. Deep learning has shown successes in a variety of computer vision tasks, where their effectiveness is dependent on the size and diversity of the image data set.In this paper, we discuss type of gaze estimation models and  related algorithms.

\section{Related Work}
\subsection{Feature based Gaze Estimation}
Feature-based methods utilizes distinctive features of the eyes like limbus and pupil contour, eye comers and cornea reflections are the common features used for gaze estimation. The aim of feature-based methods is to identify local features of the eye that are generally less sensitive to variations in illumination and viewpoint. These
systems have performance issues in outdoors or under strong ambient light.  Feature-based gaze estimation utilizes geometric vectors which map the shape of an eye and head pose to estimate gaze direction. A simple approach discussed by Sesma\cite{b3} , is the pupil-center-eye-corner vector or PC-EC vector that is used for estimating horizontal gaze direction on public displays \cite{b4}. This approach replaces corneal reflections used in eye tracking with the PC-EC vector. Another approach using Feature-based gaze model is \cite{b5} building eye gaze model using user interaction cues. Using supervised learning algorithm  to learn the correlation between gaze and interaction cues, such as cursor and caret locations.  The supervised learning algorithm would give sturdy geometric  gaze  features  and  a  ways to  identify  good  training  data  from  noisy data. This method uses behavior-informed validation\cite{b2}  to  extract  gaze  features  that  correspond  with  the interaction cue, with an average error of 4.06º.

\subsection{Model based Gaze Estimation}
 Model-based methods estimate gaze by fitting facial and eye models to the input image. As these methods use human facial features, simple parameters can describe the gaze state without a large amount of person-specific training data.Traditional model-based methods  use the shape of the iris to estimate gaze direction. \cite{b6} An ellipse is fitted to the observed iris, and then the gaze is estimated from the ellipse parameters. \cite{b7}. Downside of this approach is that it requires high-resolution eye images. Recent model-based approaches use 3D eyeball models  \cite{b7}, \cite{b10} In these approaches, the gaze direction vector is defined from the eyeball center to the iris center. The author \cite{b7} used eyelids for the eyeball model but this feature required extra annotation of the eye corners as well. The authors in \cite{b8} estimated 3D gaze vectors using tracked facial feature points; however, this requires manual annotations of eye corners and one-time calibration. The advantage of model-based methods is that they utilize the head position and rotation information obtained from the face image more productively than appearance-based methods \cite{b9}.To enhance gaze estimation combination of the head pose and eye location information is proposed, but this requires calibration phases to look at known targets. \cite{b11} One of the issues of model-based methods is the manual prior calibration process for individual users in order to estimate accurate position of the eyeballs in the face. The author \cite{b12} proposed an automatic calibration method to minimize the projected errors between model output and images through hidden online calibration. 

Model-based approaches use an explicit geometric model of the eye to estimate 3D gaze direction vector.\cite{b14} The 3D model-based gaze estimation methods use center and radius of the eyeball as well as the angular offset between visual and optical axes.The eyeball center is determined either by facial landmark such as tip of nose or by fitting deformable eye region models \cite{b13}. Most 3D model-based (or geometric) approaches use metric information from camera calibration and a global geometric model (external to the eye) of light sources, camera and monitor position and orientation. Most of the model-based method first reconstruct the optical axis of the eye, then reconstruct visual axis; finally the point of gaze is estimated by intersecting the visual axis with the scene geometry. Reconstruction of the optical axis is done by estimation of the cornea and pupil centre.

\subsection{Cross Ratio based Gaze Estimation}
In contrast to feature-based and model-based methods, cross-ratio methods achieves gaze estimation using a few IR illumination sources and the detection of their corneal reflections. \cite{b15} In this approach, suggest a new system five IR LEDs and a CCD camera are used to estimate the direction of a user's eye gaze. The IR LEDs placed on the corners of a computer monitor make glints on the cornea of the eye when user sees the monitor. It highlights the the center of a pupil creating a polygon. Without computing the geometrical relation among the eye,eye gaze can be computed from the camera, and the monitor in 3D space. Cross-ratio (CR) based methods offer many attractive properties for remote gaze estimation using a single camera in an uncalibrated setup by exploiting invariance of a plane projectivity.  To improve the performance of CR-based eye gaze trackers as the subject moves away from the calibration position, an adaptive homography mapping is used for achieving gaze prediction with higher accuracy at the calibration position and more robustness under head movements. This framework uses a learning-based method for reducing spatially-varying gaze errors and head pose dependent errors simultaneously. While these
methods are promising, additional illumination sources may not be available on unmodified devices or settings for applications such as crowd-sourced saliency estimation using commodity devices

\subsection{Appearance based Gaze Estimation}
Appearance-based methods directly use an eye image as input to estimate the point of gaze through machine learning. Previously many algorithms like adaptive linear regression \cite{b17}, support vector regression \cite{b18}, Gaussian process regression \cite{b19}, and convolutional neural networks (CNN) \cite{b8} have been proposed for point of gaze estimation. Appearance-based methods  work more effectively on low-resolution eye images than model-based methods \cite{b21}. Earlier works would use image intensities as features for linear regression,random forests and k-NN. Previously, appearance-based methods required large user-specific calibrated training datasets. Instead of calibration,  hundreds of individual samples can be used to achieve sufficient accuracy. Another approach was proposed by authors \cite{b21} who  collected the MPIIGaze dataset, which contains a large number of images of laptop users looking at on-screen markers in daily life. They trained a CNN using the dataset, then achieved person- and head pose-independent gaze estimation in the wild. But this requires high computational cost and a discrete GPU for real-time tracking.\cite{b21}
A VGG-16 network performs better than compared to the MnistNet architecture by an improvement of 0.8◦. On the architectural side, other works explore multi-modal training, such as with head pose information \cite{b21}, full-face images , or an additional “face-grid” modality for direct estimation of point of gaze \cite{b22}. Modified AlexNet used with face images \cite{b23} for the task of gaze estimation show drastic improvement in accuracy 1.9◦. CNNs learns features implicitly for personalized gaze estimation using as few as 10 calibration samples.

\cite{b2}
\section{Appearance Based Model Architecture }
\subsection{Stacked Hourglass Network }
\subsubsection{Stacked Hourglass Network Overview }

Hourglass modules are similar to auto-encoders in that feature maps are downscaled via pooling operations, then upscaled using bilinear interpolation. When given 64 feature maps, the network refines them at 4 different image scales, multiple times. This repeated bottom-up, top-down inference ensures a large effective receptive field and allows for the encoding of spatial relations between landmarks, even under occlusion.
\begin{figure}
    \includegraphics[width=0.48\textwidth]{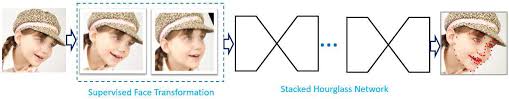}
    \caption{Stacked hour glass for facial landmark localization \cite{b24}}
    \label{shg1}
\end{figure}

\begin{figure}
    \includegraphics[width=0.48\textwidth]{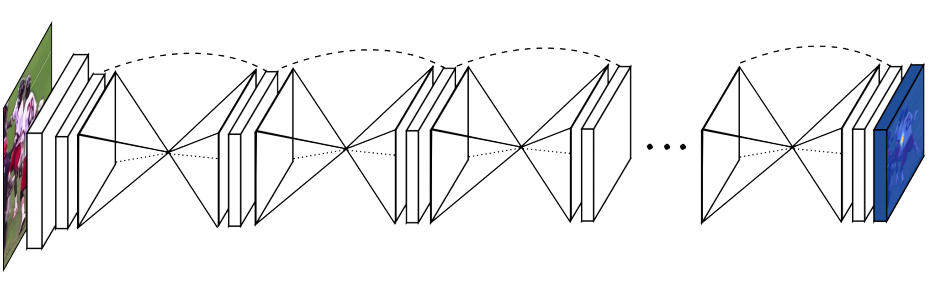}
    \caption{Stacked hourglass network \cite{b24}}
    \label{shg2}
\end{figure}

Stacked Hourglass Network (HG) is a stack of hourglass modules. It got this name because the shape of each hourglass module closely resemble an hourglass, as we can see from the picture above. The idea behind stacking multiple HG (Hourglass) modules instead of forming a giant encoder and decoder network is that each HG module will produce a full heat-map for landmark prediction. Thus, the latter HG module can learn from the landmark predictions of the previous HG module.

We use heat-map to represent facial locations in an image.This preserves the location information, and then we just need to find the peak of the heat-map.In addition, we would also calculate the loss for each intermediate prediction, which helps us to supervise not only the final output but also all HG modules effectively.

\begin{figure}
     \includegraphics[width=0.48\textwidth]{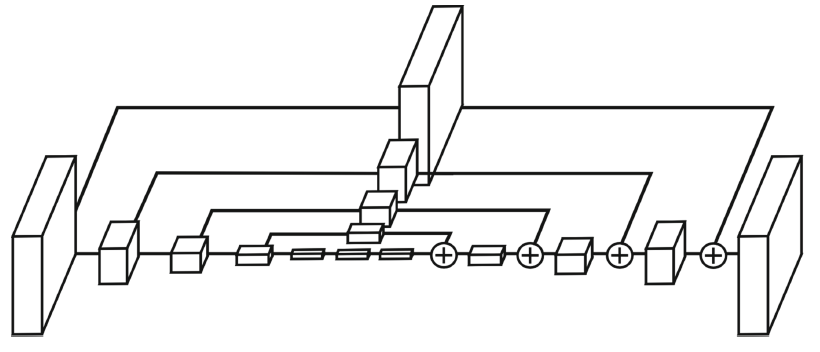}
    \caption{ An illustration of a single “hourglass” module. Each box in the figure corresponds to a residual module as seen in Figure 1. The number of features is consistent across the whole hourglass.
 \cite{b24}}
 \label{shg3}
\end{figure}
\subsubsection{Hourglass Module}

In the below diagram, each box is a residual block plus some additional operations like pooling. In general, an HG module is an encoder and decoder architecture, where we downsample the features first, and then upsample the features to recover the info and form a heat-map. Each encoder layer would have a connection to its decoder counterpart, and we could stack as many as layers we want. In the implementation, we usually make some recursions and let this HG module to repeat itself.

\subsubsection{Intermediate Supervision}

As you can see from the diagram above, when we produce something from the HG module, we split the output into two paths. The top path includes some more convolutions to further process the features and then go to the next HG module. The interesting thing happens at the bottom path. Here we use the output of that convolution layer as an intermediate heat-map result (blue box) and then calculate loss between this intermediate heat-map and the ground-truth heat-map. In other words, if we have 4 HG modules, we will need to calculate four losses in total: 3 for the intermediate result, and 1 for the final result.

\begin{figure}
    \includegraphics[width=0.48\textwidth]{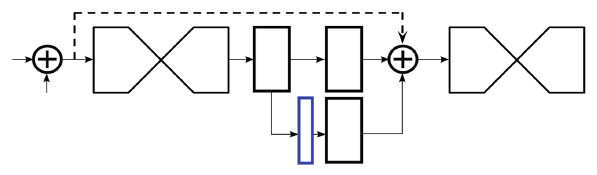}
    \caption{Intermediate supervision \cite{b24}}
    \label{shg4}
\end{figure}

\begin{figure}
    \includegraphics[width=0.48\textwidth]{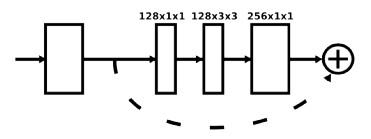}
    \caption{Residual module \cite{b24}}
    \label{shg5}
\end{figure}

\subsubsection{Loss Function}

The network performs the task of predicting heatmaps, one per eye region landmark. The heatmaps encode the per-pixel confidence on a specific landmark’s location. We place 2-dimensional Gaussians centered at the sub-pixel landmark positions such that the peak value is 1. The neural network then minimizes the l2 distance between the predicted and ground-truth heatmaps per landmark via the following loss term:

$$
\mathcal{L}_{\text {heatmaps}}=\alpha \sum_{i=1}^{18} \sum_{\mathbf{p}}\left\|\tilde{h}_{i}(\mathbf{p})-h_{i}(\mathbf{p})\right\|_{2}^{2}
$$

where h(p) is the confidence at pixel p and $\tilde{h}$ is a heatmap predicted by the network. We empirically set the weight coefficient $\alpha$  = 1.
Additionally predict an eyeball radius value $\tilde{r}_{u v}$ . This is done by first appending a soft-argmax layer [Honari et al. 2018] to calculate landmark coordinates from heatmaps, then further appending 3 linear fully-connected layers
with 100 neurons each with batch normalization and ReLU activation and one final regression layer with 1 neuron. The loss term for the eyeball radius output is:

$$
\mathcal{L}_{\text {radius}}=\beta\left\|\tilde{r}_{u v}-r_{u v}\right\|_{2}^{2}
$$

$$
\text { where we set } \beta=10^{-7} \text {and use ground-truth radius } r_{u v}
$$

\subsection{Densely Connected Convolutional Networks }
\subsubsection{Densely Connected Convolutional Networks Overview }
Densely Connected Convolutional Networks are used to deal with vanishing gradient problem about how, as networks get deeper, gradients aren’t back-propagated sufficiently to the initial layers of the network. The gradients keep getting smaller as they move backwards into the network and as a result, the initial layers lose their capacity to learn the basic low-level features.

\subsubsection{Dense connections}

Following the feed-forward nature of the network, each layer in a dense block receives feature maps from all the preceding layers, and passes its output to all subsequent layers. Feature maps received from other layers are fused through concatenation, and not through summation (like in ResNets).
\begin{figure}
    \includegraphics[width=0.48\textwidth]{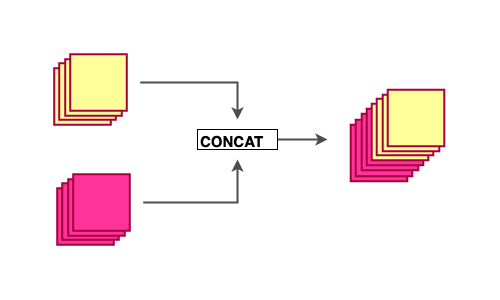}
    \caption{Concatenation of feature maps \cite{b25}}
\end{figure}
These connections form a dense circuit of pathways that allow better gradient-flow.Because of these dense connections, the model requires fewer layers, as there is no need to learn redundant feature maps, allowing the collective knowledge (features learnt collectively by the network) to be reused. The proposed architecture has narrow layers, which provide state-of-the-art results for as low as 12 channel feature maps. Fewer and narrower layers means that the model has fewer parameters to learn, making them easier to train. 
\begin{figure}
    \includegraphics[width=0.48\textwidth]{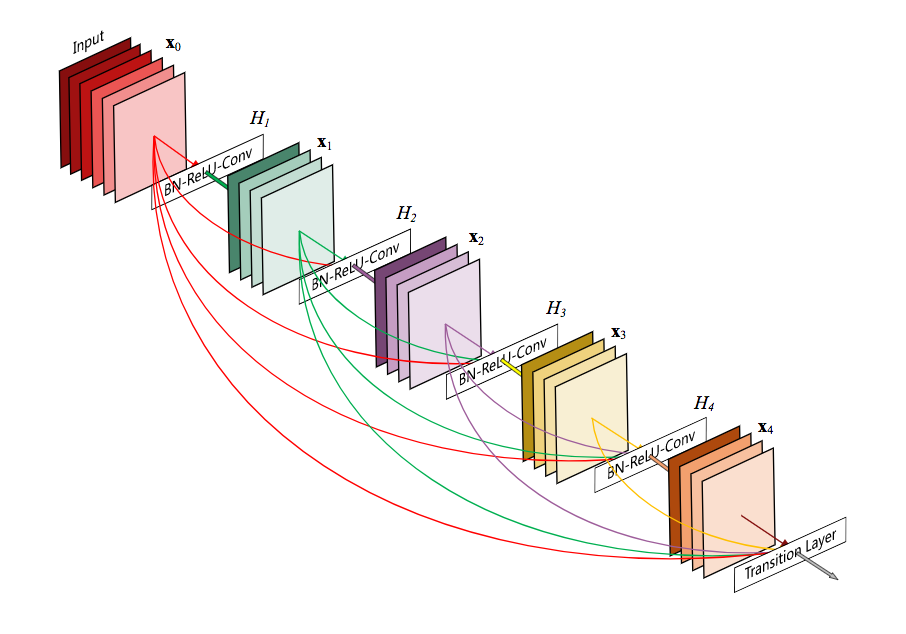}
    \caption{Each layer has direct access to the gradients of the loss function and the original input signal\cite{b25}}
\end{figure}

\subsubsection{Composite function }

\begin{figure}
    \includegraphics[width=0.48\textwidth]{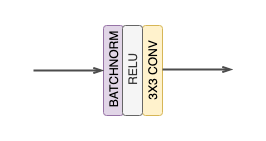}
    \caption{Composite function
 \cite{b25}}
\end{figure}
Each CONV block in the network representations in the paper[25] corresponds to an operation of BatchNorm→ReLU→Conv*

\subsubsection{Dense block}
 A dense block comprises n dense layers. These dense layers are connected using a dense circuitry such that each dense layer receives feature maps from all preceding layers and passes it’s feature maps to all subsequent layers. The dimensions of the features (width, height) stay the same in a dense block.

\begin{figure}
    \includegraphics[width=0.48\textwidth]{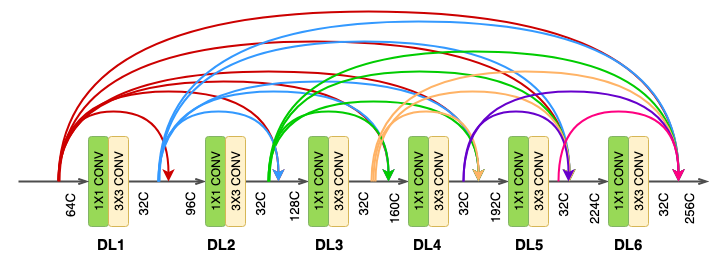}
    \caption{Dense block with channel count (C) of features entering and exiting the layers \cite{b25}}
\end{figure}

Dense layer

Each dense-layer consists of 2 convolutional operations -

1 X 1 CONV (conventional conv operation for extracting features)

3 X 3 CONV (bringing down the feature depth/channel count)

The DenseNet-121 comprises of 6 such dense layers in a dense block. The depth of the output of each dense-layer is equal to the growth rate of the dense block.

\subsubsection{Transition layer
}

A transition layer (or block) is added between two dense blocks. The transition layer consists of -

1 X 1 CONV operation

2 X 2 AVG POOL operation

The 1 X 1 CONV operation reduces the channel count to half.

The 2 X 2 AVG POOL layer is responsible for downsampling the features in terms of the width and height.

\begin{figure}
    \includegraphics[width=0.48\textwidth]{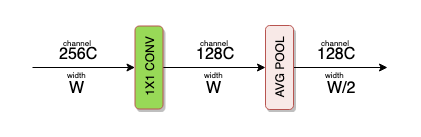}
    \caption{Transition layer \cite{b25}}
\end{figure}

\subsubsection{Full network}

Different number of dense layers for each of the three dense block.
\begin{figure}
    \includegraphics[width=0.48\textwidth]{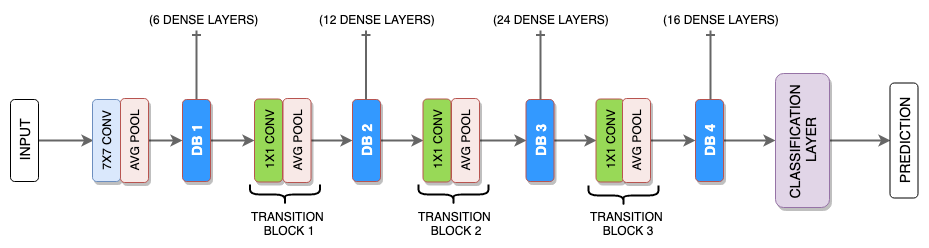}
    \caption{Full network
 \cite{b25}}
\end{figure}

\subsubsection{Loss Function}

This approach models input image into an intermediate image representation of the eye as gazemap i.e. $\bold{m}$. Gaze direction $\bold{g}$ is defined as :
$$
\boldsymbol{g}=k \circ j(x) \text { where } j: \boldsymbol{x} \rightarrow \boldsymbol{m} \text { and } k: \boldsymbol{m} \rightarrow \boldsymbol{g}
$$

So using gamzemap m, we can estimate gaze direction g. Considering a simple model of the human eyeball and iris where iris diameter is approximately 12mm and eyeball diameter is approximately 24mm, we calculate the iris centre coordinates $\left( u_{i} , v_{i} \right) $:
$$
\begin{array}{l}
u_{i}=\frac{m}{2}-r^{\prime} \sin \phi \cos \theta \\
v_{i}=\frac{n}{2}-r^{\prime} \sin \theta
\end{array}
$$

where $
r^{\prime}=r \cos \left(\sin ^{-1} \frac{1}{2}\right)
$
, and gaze direction $
\boldsymbol{g}=(\theta, \phi)
$. For regression to g, using DenseNet architecture explained above, we perform image classification and gain desired output confidence maps via 1 x 1 convolutional layers. The loss term for gaze direction is :

$$
\mathcal{L}_{\text {gaze }}=\|\boldsymbol{g}-\hat{\boldsymbol{g}}\|_{2}^{2}
$$

where $\hat{\boldsymbol{g}}$ is the gaze direction predicted by out network.

In this gazemap network, we use 3 hourglass modules with intermediate supervision applied on the gazemap outputs of the last module only. The minimized intermediate loss is:

$$
\mathcal{L}_{\text {gazemap }}=-\alpha \sum_{p \in \mathcal{P}} \boldsymbol{m}(p) \log \hat{\boldsymbol{m}}(p)
$$
where we calculate a cross-entropy between predicted $\hat{\boldsymbol{m}}$ and ground-truth gazemap m for pixels p in set of all pixels P. In our evaluations, we set the coefficient $\alpha$ to $10^{ - 5}$.

\section{Experimental Analysis}
\subsection{Hyperparameters}
Hyperparameters  modified and tested include  -  L2 weights regularization coefficient, batch size, Alpha: learning rate, number of epochs. For the parameters - Number of layers, Size of layers, Activation function, Learning rate, ReLU activation function was used. Slight data augmentation is applied in terms of image translation and scaling and learning rate is multiplied by 0.1 after every 5k gradient
update steps, to address over-fitting and to stabilize the final error. Number of stacks in hourglass network is increased to train on 8 hourglass modules with intermediate supervision yields significantly improved landmark localization accuracy compared to 2-stack or 4-stack models with the same number of model parameters.

\section{Results Analysis}
 \subsection{ Error Analysis}

The result show the mean angular error (MAE) of vectors of pitch and yaw for different parameters and models. We mainly discuss two models : Gazemap approach (Model 1) using dense net architecture with stacked hourglass models and  Eye region landmark approach  (Model 2) using identification of eye region and iris localization landmarks as input for the stacked hourglass network. Below models were trained on 150k entries.  
  
Different parameters for Model 1:Gazemap approach. Mean angular error is in degrees.

Different number of dense block modules and different umber of layers per block:
\begin{center}
\begin{tabular}{|l|l|l|}
    \hline Paramaters  & Gazemap MAE  \\
    \hline 5 dense blocks(5 layers) & 5.88  \\
    \hline 5 dense blocks(3 layers) &  9.09  \\
    \hline 5 dense blocks(6 layers) &  4.6\\
    \hline 6 dense blocks(5 layers) &  4.5  \\
    \hline 4 dense blocks(5 layers) &  7.6\\
    \hline
\end{tabular}
\end{center}

Different learning parameters:
\begin{center}
\begin{tabular}{|l|l|l|}
    \hline Paramaters  & Gazemap MAE  \\
    \hline alpha = 0.001 & 6.78 \\
    \hline alpha = 0.0001 & 4.65   \\
    \hline alpha = 0.00001 & 4.6 \\
    \hline
\end{tabular}
\end{center} 
Different parameters for Model 2:Eye region landmark approach. Mean angular error is in degrees.

Different number of stacked modules:

\begin{center}
\begin{tabular}{|l|l|l|}
    \hline Paramaters  & Model 2 MAE  \\
    \hline 3 stacked hourglass modules & 5.3  \\
    \hline 8 stacked hourglass modules &  4.09  \\
    \hline 2 stacked hourglass modules &  7.3\\
    \hline
\end{tabular}
\end{center}

Different learning parameters:
\begin{center}
\begin{tabular}{|l|l|l|}
    \hline Paramaters  & Model 2 MAE  \\
    \hline alpha = 0.001 & 8.5 \\
    \hline alpha = 0.0001 & 4.9   \\
    \hline alpha = 0.00001 & 4.6 \\
    \hline
\end{tabular}
\end{center}
Changing batchsize did not affect the accuracy whereas increasing the number of epochs ($< 100$) did not affect the results. The above graph shows above mean angular error for pitch and yaw angles.
\begin{figure}
    \includegraphics[width=0.48\textwidth]{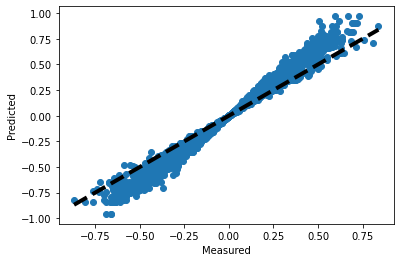}
    \caption{Predicted vs Actual gaze estimation angle pitch}
\end{figure}
\begin{figure}
    \includegraphics[width=0.48\textwidth]{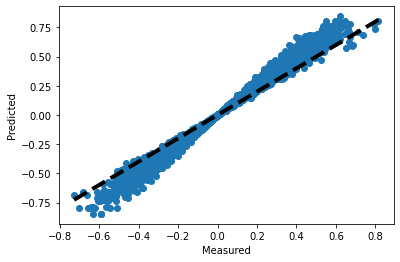}
    \caption{Predicted vs Actual gaze estimation angle yaw}
\end{figure}

\begin{figure}
    \includegraphics[width=0.48\textwidth]{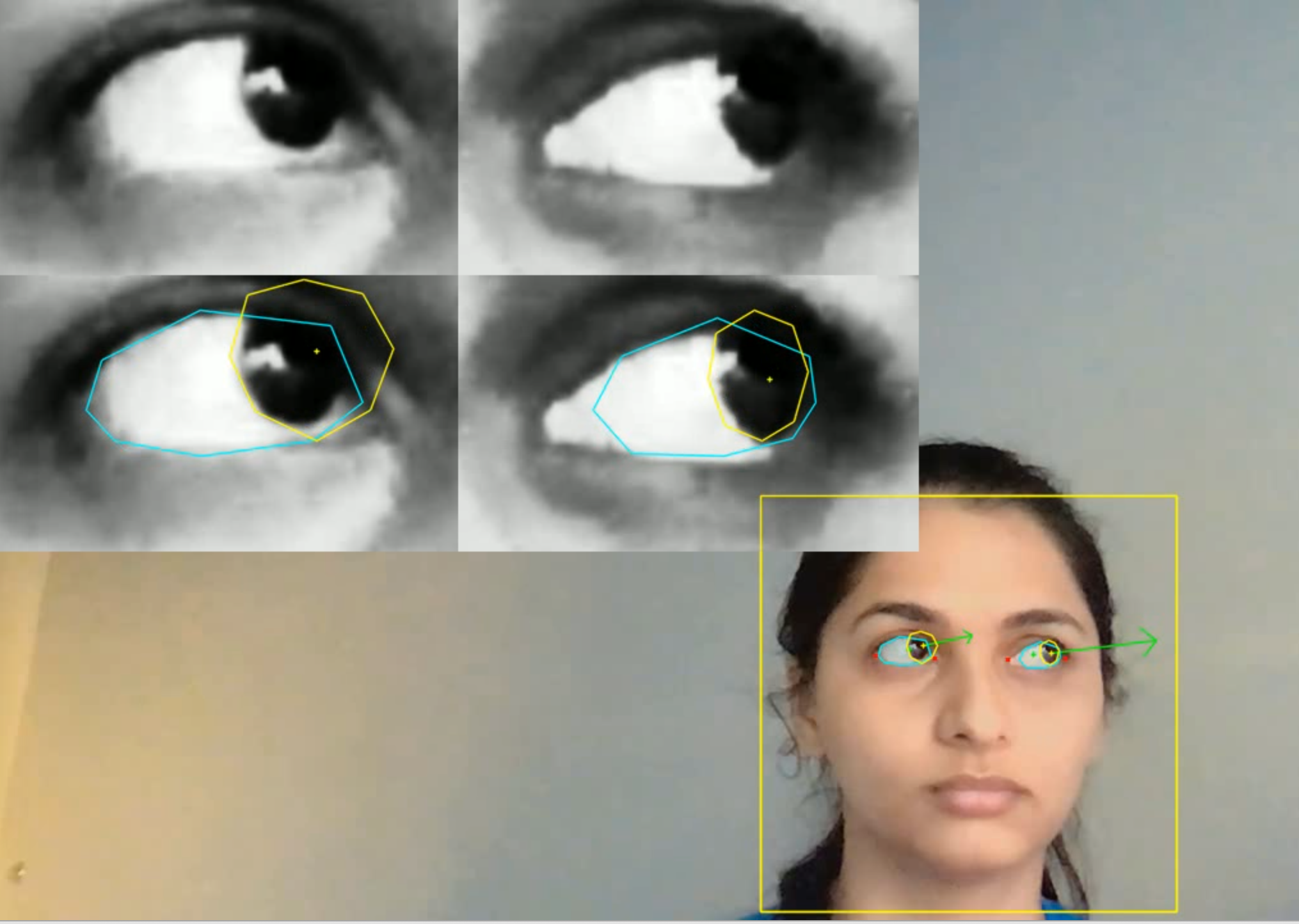}
    \caption{Demo application}
    \label{demo}
\end{figure}

Figure~\ref{demo} is the demo image of running the model on a web-camera video file.
\subsection{Future Work}\label{SCM}
There are many different approaches to identify different eye region landmarks and try different representations of landmarks to gain more robust gaze estimation model. This model can be easily adapted to work on low resolution data sets. The current work is made open source and available on https://github.com/aveenakottwani/EyeGazeEstimationModels. 

\vspace{12pt}
\color{red}

\end{document}